\documentclass[10pt,a4paper,fleqn]{article}
\usepackage[english]{babel}
\usepackage[none]{hyphenat}
\usepackage{newtxtext,newtxmath}

\usepackage[left=2.5cm,right=2.5cm,top=2.5cm,bottom=2.5cm]{geometry}
\usepackage[compact]{titlesec}
\titleformat{\section}{\sc\normalsize\bfseries}{\thesection}{1em}{}

\usepackage{graphicx}
\graphicspath{{./}{./figures/}}
\usepackage{ccaption}
\usepackage{xcolor}
\usepackage{xspace}
\usepackage{hyperref}
\hypersetup{pdfborderstyle={/S/S/W 1},colorlinks=true,linkcolor=blue,citecolor=blue,filecolor=blue,urlcolor=blue}
\hypersetup{pdfborder={0 0 0},citebordercolor={0 0 0},filebordercolor={0 0 0},linkbordercolor={0 0 0},menubordercolor={0 0 0},urlbordercolor={0 0 0}}
\hypersetup{pdftitle={Opportunistic Emulation of Computationally Expensive Simulations via Deep Learning}}
\hypersetup{pdfauthor={Conrad Sanderson, Dan Pagendam, Brendan Power, Frederick Bennett, Ross Darnell}}

\makeatletter
\renewcommand{\fnum@figure}[1]{\textbf{\figurename~\thefigure}.~}
\renewcommand{\fnum@table}[1]{\textbf{\tablename~\thetable}.~}
\makeatother

\def\eg{eg.\xspace}
\def\ie{ie.\xspace}

\textfloatsep=\floatsep           
\intextsep=\floatsep              

\makeatletter
\renewcommand\@makefntext[1]{%
\setlength\parindent{1em}%
\noindent
{#1}}
\makeatother

\makeatletter
\def\blfootnote{\xdef\@thefnmark{}\@footnotetext}
\makeatother

\begin{document}

\begin{center}
\textbf{\LARGE Opportunistic Emulation of Computationally Expensive Simulations via Deep Learning}

~\\

\textbf{Conrad Sanderson}{\tiny~}\textsuperscript{\textdagger\textdaggerdbl},
\textbf{Dan Pagendam}{\tiny~}\textsuperscript{\textdagger},
\textbf{Brendan Power}{\tiny~}\textsuperscript{$\bigtriangleup$},
\textbf{Frederick Bennett}{\tiny~}\textsuperscript{$\diamond$},
\textbf{Ross Darnell}{\tiny~}\textsuperscript{\textdagger}

~\\

\textsuperscript{\textdagger}{\tiny~}\textit{Data61~/~CSIRO, Australia}\\
\textsuperscript{\textdaggerdbl}{\tiny~}\textit{Griffith University, Australia}\\
\textsuperscript{$\bigtriangleup$}{\tiny~}\textit{Department of Resources, Queensland Government, Australia}\\
\textsuperscript{$\diamond$}{\tiny~}\textit{Department of Environment and Science, Queensland Government, Australia}\\

\end{center}

\vspace{1ex}
\section*{Abstract}
\vspace{-1ex}

With the underlying aim of increasing efficiency of computational modelling pertinent for managing and protecting the Great Barrier Reef,
we perform a preliminary investigation on the use of deep neural networks
for {\it opportunistic} model emulation of APSIM models by repurposing an existing large dataset containing the outputs of APSIM model runs.
The dataset has not been specifically tailored for the model emulation task.
We employ two neural network architectures for the emulation task:
densely connected feed-forward neural network (FFNN),
and gated recurrent unit feeding into FFNN (GRU-FFNN), a type of a recurrent neural network.
Various configurations of the architectures are trialled.
A minimum correlation statistic is employed to identify clusters of APSIM scenarios that can be aggregated to form training sets for model emulation.

We focus on emulating four important outputs of the APSIM model:
{(i)} {\it runoff} -- the amount of water removed as runoff,
{(ii)} {\it soil\_loss} -- the amount of soil lost via erosion,
{(iii)} {\it DINrunoff} -- the mass of dissolved inorganic nitrogen exported in runoff,
and
{(iv)} {\it Nleached} -- the mass of nitrogen leached in water draining to the water table.
The GRU-FFNN architecture with three hidden layers and 128 units per layer provides good emulation of {\it runoff} and {\it DINrunoff}.
However, {\it soil\_loss} and {\it Nleached} were emulated relatively poorly under a wide range of the considered architectures;
the emulators failed to capture variability at higher values of these two outputs.

While the opportunistic data available from past modelling activities provides a large and useful dataset for exploring APSIM emulation,
it may not be sufficiently rich enough for successful deep learning of more complex model dynamics.
Design of Computer Experiments may be required to generate more informative data to emulate all output variables of interest.
We also suggest the use of synthetic meteorology settings to allow the model to be fed a wide range of inputs.
These need not all be representative of normal conditions,
but can provide a denser, more informative dataset from which complex relationships between input and outputs can be learned.
\blfootnote{\textbf{Published in:} International Congress on Modelling and Simulation, 2021. DOI: 10.36334/modsim.2021.L1.sanderson}

\vspace{1ex}

\textit{\textbf{Keywords:} surrogate models, model emulation, deep neural networks, recurrent neural networks, APSIM}

\vspace{1ex}
\section{Introduction}
\vspace{-1ex}

Model emulation is a technique by which a computationally burdensome (usually physically motivated) model is replaced with a computationally efficient surrogate~\cite{asher_2015}.
Model emulation first gathered traction in the statistical literature with the work of Kennedy and O'Hagan~\cite{kennedy_2001} 
who advocated the use of Gaussian Processes (GPs) for the surrogate model.
Since then, other statistical and machine learning models have also been successfully used as model emulators to overcome deficiencies of GPs.
In particular, GPs lose computational efficiency when the dataset used to train the emulator is large and/or the outputs of the model are high-dimensional.
Both of these issues are common when dealing with complex, physically motivated environmental models.

To overcome the abovementioned problems, practitioners in recent years have advocated the use of other machine learning approaches
such as Random Forests \cite{gladish_2018, hooten_2011, leeds_2013}
and Deep Neural Networks (DNNs) \cite{Kasim_arxiv_2020, pal_2019, Puscasu_2014, Sit_WST_2020}.
Standard implementations of Random Forests suffer from the problem that they only predict univariate outputs.
For this reason, their use requires an additional step of dimension reduction which typically results in reduced emulator accuracy.
In contrast, DNNs offer a method that can handle high-dimensional inputs (parameters, forcing variables) and model outputs.
In addition, use of DNNs has been democratised via the availability of sophisticated open-source computational tools,
such as TensorFlow \cite{TensorFlow_arXiv_2016}, PyTorch \cite{PyTorch_NeurIPS_2019} and mlpack \cite{mlpack2018},
which can make use of Graphics Processing Units (GPUs) for accelerated computation.

\newpage

As neural network models are relatively straightforward and are essentially
described by a large set of numerical weight parameters,
surrogate models based on neural networks have additional useful characteristics
besides providing potential computational savings.
Such models are:
{\bf (i)}~portable across computing architectures (\eg~CPUs, GPUs, operating systems),
{\bf (ii)}~relatively easily deployed in cloud-based environments,
{\bf (iii)}~relatively easily integrated into larger frameworks
(\eg~using the outputs of surrogate models as inputs into other systems).

The Science Division within the Queensland Government Department of Environment and Science
relies on the use of model driven simulations 
for the Paddock to Reef (P2R) Integrated Monitoring, Modelling and Reporting Program,
which focuses on tracking water quality and associated catchment systems.
The models are used for evaluating many ``what-if'' scenarios,
which may require millions of simulations (model~runs).

Currently, the Agricultural Production Systems sIMulator (APSIM)~\cite{Holzworth_2014}
and HowLeaky~\cite{McClymont_2011,Shaw_2011} models are used for P2R.
These models have been developed and calibrated using measured data.
They provide critical information on surface runoff, the movement of sediment, nutrients and pesticides,
for the range of land use, land management practices, climate regimes, and soil types
across Great Barrier Reef (GBR) catchments.
The time series outputs from these models are used for the {\it GBR Report Card},
as part of a larger framework for managing and protecting the~GBR.
The complexity of the models, combined with the millions of runs,
results in considerable computational time and labour effort, sometimes lasting several months.
Furthermore, not all the outputs generated by the models are required to meet the objectives for the {\it GBR Report Card}.

A simplification of the original models through model emulation
may achieve lower computational requirements while retaining accuracy within acceptable bounds.
This can lead to faster execution (thereby freeing resources for other tasks such as higher level analysis and interpretation),
and/or increasing the number of model runs within a given time budget.

An important aspect of building a model emulator
is the design of computer experiments for generating data (model inputs and outputs)
that will be informative for training emulators~\cite{santner_2003}.
However, there are situations where it may not be possible to accomplish this,
due to the limitations imposed by a given pre-made dataset,
and/or infeasibility of creating a new dataset specifically tailored for the model emulation task
due to lack of resources (\eg lack of time, personnel, and/or prolonged access to shared computing hardware).

In this work we undertake a preliminary investigation of {\it opportunistic} model emulation
by repurposing an existing large dataset (outputs of APSIM model runs) previously used for the {\it GBR Report Card},
that has not been specifically tailored for model emulation.
We examine whether important water quality constituents
that are outputs of the P2R cane farming APSIM model can be reliably emulated using DNNs
within the constraints of the available dataset.

We continue the paper as follows.
Section~\ref{sec:data_overview} briefly describes the pre-existing dataset of APSIM model runs,
the variables to be emulated, and the generation of feature vectors with the aim of facilitating emulation.
Section~\ref{sec:model_emulation} overviews two neural network architectures used for emulation in this work.
Section~\ref{sec:experiments} provides an empirical evaluation of the two architectures.
The main findings and future avenues are discussed in Section~\ref{sec:conclusion}.

\vspace{1ex}
\section{Dataset \& Preparation}
\label{sec:data_overview}
\vspace{-1ex}

In brief, the available P2R dataset of APSIM model runs is comprised of a large set of time series covering many variables.
Each model run is for a specific combination of several forcing variables, including:
soil~type, farm management class, meteorology (\eg ~rain patterns), simulation start year, and degree of soil hydraulic conductivity.
The data for each model run has 35 columns, covering both forcing and output variables of interest;
each row represents the state of the simulation for one day.
As each run represents a simulation over many decades, there is typically around 18,000 time-steps (number of rows).

While the dataset is large and spans a wide range of management scenarios,
there is no guarantee that the dataset is fit-for-purpose for building emulators.
It should be noted that the time series contain a very large number of zeros
(long periods or dormancy in output variables between significant rainfall events),
which may limit the ability of the emulator to learn system dynamics.

To ameliorate the above issue, feature vectors (predictor variables) derived from APSIM data
were generated in order to potentially make the process of emulating model dynamics easier.
These variables are outlined in Table~\ref{em:inputVariables} and were used to
{\bf (i)}~code the specific management practices using numeric variables,
and
{\bf (ii)}~provide long-range temporal context that can allow past agricultural management and meteorological conditions to be reflected in temporally localised predictions.

\newpage
We focus on emulating four important outputs of the APSIM model, namely:
{\bf (i)} {\it runoff} -- the amount of water removed as runoff on each day,
{\bf (ii)} {\it soil\_loss} -- the amount of soil lost via erosion on each day,
{\bf (iii)} {\it DINrunoff} -- the mass of dissolved inorganic nitrogen exported in runoff on each day,
and
{\bf (iv)} {\it Nleached} -- the mass of nitrogen leached in water draining to the water table.

Prior to training emulators, all of the input variables listed in Table~\ref{em:inputVariables} and output variables were scaled as
\mbox{\small $x^{*}_{i, k} = \left\{ x_{i, k} - \min_{s \in S_1}(x_{s, k}) \right\} / \left\{ \max_{s \in S_1}(x_{s, k}) - \min_{s \in S_1}(x_{s, k}) \right\}$},
where $x_{i, k}$ is the $i$-th record of the $k$-th variable and $S_1$ is set of all training data.
Scaling the inputs and sometimes the outputs of neural networks is common practice~\cite{ensmallen_JMLR_2021,Goodfellow_2016},
and ensures that gradient-based optimisation methods
used for updating the weights of neural network emulators are numerically stable.

\begin{table}[!t]
\centering
\small
\begin{tabular}{|c|c|}
\hline
\bf{Generated Feature Name} & \bf{Description}  \\ \hline\hline
Management 1 & Numeric encoding of soil management  \\ \hline
Management 2 & Numeric encoding of pesticide management  \\ \hline
Management 3 & Numeric encoding of fertiliser management  \\ \hline
Management 4  & Numeric encoding of millmud use  \\ \hline
Conductivity  & Numeric encoding soil conductivity \\ \hline 
Day of Year & Numeric encoding of day of year \\ \hline
Seasonal & Trigonometric variables encoding seasonality \\ \hline
Planting Month & Numeric encoding if this was a planting month \\ \hline
Planting Year & Numeric encoding if this was a planting year \\ \hline
Time Since Cane Planted & Number of days since cane planted \\ \hline 
Time Since Cowpea Planted & Number of days since cowpea planted \\ \hline
Time Since Fertiliser Applied & Number of days since fertiliser applied \\ \hline
Time Since Millmud Applied & Number of days since millmud applied \\ \hline
Cane in & Indicator for whether cane is in ground \\ \hline
Cowpea in & Indicator for whether cowpea is in ground \\ \hline 
Persistent Fertiliser & Previous (nonzero) fertiliser amount applied \\ \hline
Persistent Millmud & Previous (nonzero) millmud amount applied \\ \hline
Rainfall & Daily rainfall \\ \hline
Evaporation & Daily evaporation \\ \hline
Fertiliser & Daily fertiliser \\ \hline
Millmud & Daily millmud \\ \hline
Smoothed Rainfall & Number of exponentially smoothed rainfall series \\ \hline
Smoothed Evaporation & Number of exponentially smoothed evaporation series \\ \hline
Smoothed Fertiliser & Number of exponentially smoothed fertiliser series \\ \hline 
Smoothed Millmud & Number of exponentially smoothed millmud series \\ \hline
\end{tabular}
\caption{Features derived from APSIM data and used for emulator construction.}
\label{em:inputVariables}
\end{table}%

\vspace{1ex}
\section{Model Emulation}
\label{sec:model_emulation}
\vspace{-1ex}

\begin{figure}[!b]
\begin{minipage}{1\textwidth}
  \centering
  \begin{minipage}{0.37\textwidth}
    \vspace{7ex}
    \includegraphics[width=1\textwidth,height=1.096\textwidth]{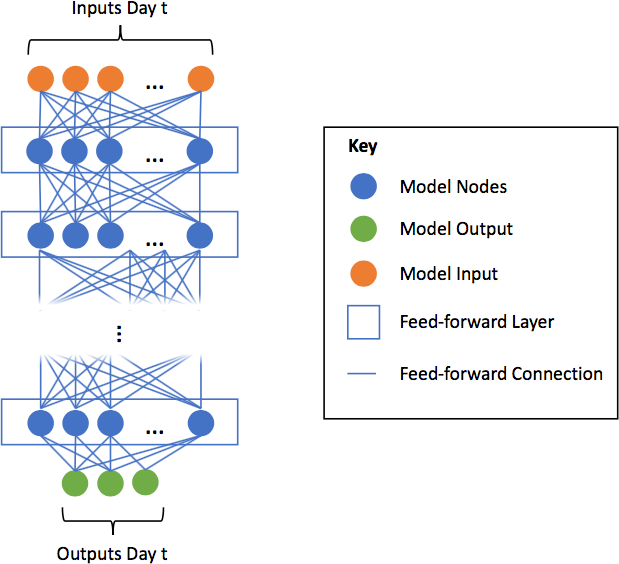}
    \vspace{8ex}
    \caption{\small Example of a Feed-Forward Neural Network (FFNN) for model emulation.}
    \label{em:ffnn-diag}
  \end{minipage}
  \hfill
  \vline
  \vline
  \hfill
  \begin{minipage}{0.56\textwidth}
    \includegraphics[width=1\textwidth,height=1.096\textwidth]{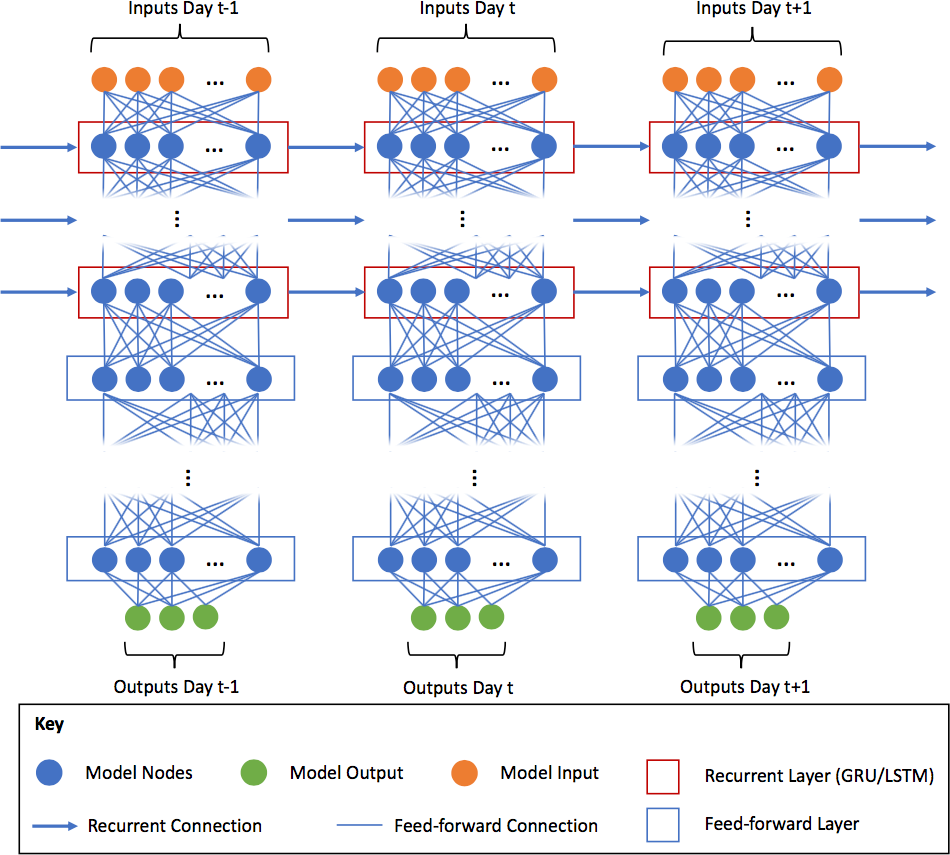}
    \caption{\scalebox{0.86}{Example of a recurrent neural network feeding into FFNN.}}
  \label{em:rnn-diag}
  \end{minipage}
\end{minipage}
\end{figure}

In this work we explore two DNN-based emulators,
namely densely connected feed-forward neural networks (FFNNs),
and one type of a recurrent neural network: gated recurrent unit feeding into FFNN (GRU-FFNN).
Conceptual examples are shown in Figures \ref{em:ffnn-diag} and \ref{em:rnn-diag}.
Recurrent neural networks have been previously used for sequence processing
and may have better ability in capturing the underlying system dynamics.
While not considered here, other types of recurrent neural networks are possible,
such as Long Short-Term Memory (LSTM) networks~\cite{Goodfellow_2016}.
In essence, each emulator is trained by using known values of the variables of interest
({\it runoff}, {\it soil\_loss}, {\it DINrunoff}, {\it Nleached})
in conjunction with the features shown in Table~\ref{em:inputVariables}, derived from APSIM data.

A FFNN with $K$ layers is a deterministic model that maps a vector of inputs at time~$t$, denoted $\mathbf{x}_t$, (of length~$d$)
to a vector of model outputs at time $t$, $\mathbf{y}_K$,  (of length $l_k$).
This mapping is performed through a number of intermediate vectors $\mathbf{y}_{k,t}$ $(k = 2, .., K-1)$ (each of length~$l_k$).
These intermediate vectors represent hidden states or layers in the FFNN as shown in Figure~\ref{em:ffnn-diag};
each of them is simply computed as
{\small $\mathbf{y}_{1,t} = h(\mathbf{W}_1 \mathbf{x}_t + \mathbf{b}_1)$}
and
{\small $\mathbf{y}_{k,t} = h(\mathbf{W}_k \mathbf{y}_{k - 1, t} + \mathbf{b}_k),  (k = 2, ..., K)$},
where
{\small $\mathbf{W}_1$} is an {\small $l_1 \times d$} matrix of weight parameters,
{\small $\mathbf{W}_k$ ($k = 2, ..., K$)} is an {\small $l_{k} \times l_{k-1}$} matrix of weight parameters,
and
{\small $\mathbf{b}_k$ ($k = 1, ..., K$)} is a length {\small $l_k$} vector of bias parameters for layer~{\small $k$}.
The function $h(\cdot)$ is referred to as an \emph{activation function};
we use the \emph{Rectified Linear Unit} function defined as
{\small $h(\mathbf{y}) = (\max(0, y_1), \max(0, y_2), ..., \max(0, y_n))^{\top}$}.

\newpage
The recurrent neural network architectures considered here can be described as having $J$ recurrent layers followed by $K$ feed-forward layers.
For the GRU-FFNN emulator, this mapping from input vector $\mathbf{x}$ to outputs $\mathbf{y}_{J + K}$ at time~$t$, can be written as

\noindent
\begin{eqnarray*}
\mathbf{z}_{1, t} =& \sigma_{g}(\mathbf{W}_{z, 1} \mathbf{x}_{t} + \mathbf{U}_{z, 1}\mathbf{h}_{1, t-1} + \mathbf{b}_{z, 1}) \\
\mathbf{r}_{1, t} =& \sigma_{g}(\mathbf{W}_{r, 1} \mathbf{x}_{t} + \mathbf{U}_{r, 1}\mathbf{h}_{1, t-1} + \mathbf{b}_{r, 1}) \\
\widehat{\mathbf{h}}_{1, t} =& \sigma_{h}(\mathbf{W}_{h, 1} \mathbf{x}_{t} + \mathbf{U}_{h, 1}\mathbf{h}_{1, t-1} + \mathbf{b}_{h, 1}) \\
~\\
\mathbf{z}_{j, t} =& \sigma_{g}(\mathbf{W}_{z, j} \mathbf{h}_{j-1, t} + \mathbf{U}_{z, j}\mathbf{h}_{j, t-1} + \mathbf{b}_{z, j})& (j = 2, ..., J) \\
\mathbf{r}_{j, t} =& \sigma_{g}(\mathbf{W}_{r, j} \mathbf{h}_{j-1, t} + \mathbf{U}_{r, j}\mathbf{h}_{j, t-1} + \mathbf{b}_{r, j}) & (j = 2, ..., J) \\
\widehat{\mathbf{h}}_{j, t} =& \sigma_{h}(\mathbf{W}_{h, j} \mathbf{h}_{j-1, t} + \mathbf{U}_{h, j}\mathbf{h}_{j, t-1} + \mathbf{b}_{h, j}) & (j = 2, ..., J) \\
~\\
\mathbf{h}_{j, t} =& (1 - \mathbf{z}_{j, t}) \circ \mathbf{h}_{j, t-1} + \mathbf{z}_{j, t} \circ \hat{\mathbf{h}}_{j, t} & (j = 2, ..., J) \\
~\\
\mathbf{y}_{1} =& h(\mathbf{W}_1 \mathbf{h}_{J, t} + \mathbf{b}_1)  \\
\mathbf{y}_k   =& h(\mathbf{W}_k \mathbf{y}_{k - 1} + \mathbf{b}_k) & (k = 2, ..., K)
\end{eqnarray*}%

\noindent
where
$\circ$ is the Hadamard (element-wise) product of two matrices/vectors,
$\mathbf{x}_{t} \in \mathbb{R}^d$ is the vector of inputs (see Table~\ref{em:inputVariables}) for day~$t$,
{\small $\mathbf{z}_{j, t}  \in \mathbb{R}^h$} is the update-gate activation vector for the $j$-th recurrent layer on day $t$,
{\small $\mathbf{r}_{j, t}  \in \mathbb{R}^h$} is the update-gate activation vector for the $j$-th recurrent layer on day $t$,
{\small $\widehat{\mathbf{h}}_{j, t}  \in \mathbb{R}^h$} is the candidate activation vector for the $j$-th recurrent layer on day $t$
and
{\small $\mathbf{h}_{j, t} \in \mathbb{R}^h$} is the hidden state vector.
$\mathbf{W}$~and $\mathbf{U}$ are matrices of weight parameters, while $\mathbf{b}$ are vectors of bias parameters.
Each matrix $\mathbf{W}$ has the size of $h \times d$,
each matrix $\mathbf{U}$ has the size of $h \times h$,
and
each vector $\mathbf{b}$ has the size of $h \times 1$.

The activation functions used in the recurrent layers are sigmoid for~$\sigma_g$, and hyperbolic tangent functions for~$\sigma_h$. 
The $K$ feed-forward layers corresponding to the vectors $\mathbf{y}_{1}, ..., \mathbf{y}_{K}$ are as defined for the FFNN model defined above.

\section{Experiments}
\label{sec:experiments}
\vspace{-1ex}

Since DNNs typically require large amounts of data to obtain reliable models~\cite{Goodfellow_2016},
we first perform a correlation analysis to ascertain which model runs are sufficiently similar to each other,
so that they can be pooled to jointly learn model dynamics.
In essence the correlation analysis is used to create clusters of similar model runs.  

We compare the outputs of multiple model runs
where each model run was conditioned on particular farm management practices,
soil types and meteorological outputs from a set of possible scenarios.
Each model run also had multiple model outputs and any type of clustering would require that all of these outputs produced sufficiently similar time series.
In order to score the level of similarity between model runs,
we chose to use a conservative approach that would group two model runs according to the minimum correlation between time series,
where the minimum was taken over the individual outputs ({\it runoff}, {\it soil\_loss}, {\it DINrunoff}, {\it Nleached}).
Specifically, we calculated this minimum correlation as
$C_{\textup{min}}(i, j) = \min_{k} \textup{Corr}(\mathbf{x}_{i, k}, \mathbf{x}_{j, k})$,
where $\mathbf{x}_{i, k}$ is the $k$-th model output (\eg soil loss) for model run $i$,
while $\textup{Corr}(\cdot)$ represents the sample correlation computed over two vectors.

The correlation function was used to populate symmetric, minimum correlation matrices that can be used to identify clusters.
Correlation matrices were produced
for each unique combination of meteorology settings, soil type, soil permeability, planting month and planting year,
while allowing the management scenario variable to vary.
For each correlation matrix, it was then possible to iterate over each row and each column
to identify similar model runs across management scenarios that exceeded a minimum correlation threshold (say 0.95).

For both emulation architectures (FFNN and GRU-FFNN),
we trialled a range of combinations of hyper-parameters as outlined in Table~\ref{em:designParams}.
For the feed-forward parts of the network architecture,
we used a funnel shape to the network, whereby the number of nodes in the network decreased with increasing depth.
We achieved this by reducing the number of nodes by a half with each layer of depth,
starting with the number of nodes listed in Table~\ref{em:designParams}.
This conical shape is commonly implemented in FFNN architectures as it helps to reduce the number of parameters/weights in a deep network
and can help to facilitate learning that is generalisable rather than simple memorisation of the training data.

\begin{table}[!b]
\centering
\begin{tabular}{|c|c|}
\hline
\bf{Architecture Design Choice} & \bf{Values Considered}  \\ \hline\hline
Model Architecture & FFNN, GRU-FFNN  \\ \hline
Num.~of Lags (days) for Training & 7, 14, 28  \\ \hline
Num.~of Hidden Recurrent Layers & 1, 3, 5, 7, 9  \\ \hline
Num.~of Hidden Feed-Forward Layers & 1, 3, 5, 7, 9  \\ \hline
Num.~of Nodes (Recurrent Layers) & 32, 64, 128, 256, 512  \\ \hline
Num.~of Nodes (Feed-Forward Layers) & 32, 64, 128, 256, 512, 1024  \\ \hline
\end{tabular}
\caption{Architecture design parameters for DNN based emulators of APSIM outputs.}
\label{em:designParams}
\end{table}%

For training the models,
we divided the model runs into three independent,
non-overlapping periods in the time series used for:
{\bf (i)} model training (1~Jan~1970 to 31~Dec~2010),
{\bf (ii)} model validation (1~Jan~2011 to 31~Dec~2015),
and
{\bf (iii)} model testing (1~Jan~2016 to 19~Nov~2018).
The training data was used for training the models in Tensorflow~\cite{TensorFlow_arXiv_2016} via Keras~\cite{chollet2015keras},
using the mini-batch Stochastic Gradient Descent algorithm \cite{dekel_2012}.
Training was performed using mini-batches of 128 randomly selected records at a time.
In essence, training data was used for updating the weights of the neural network models;
validation data was used for assessing overfitting and early stopping;
and test data was used for emulator model selection.

\begin{figure}[!t]
\begin{minipage}{1\textwidth}
  \centering
  \begin{minipage}{1\textwidth}
  \centering
  \footnotesize
  {\bf Emulation via FFNN}\\
  \includegraphics[height=0.65\textwidth,width=0.78\textwidth]{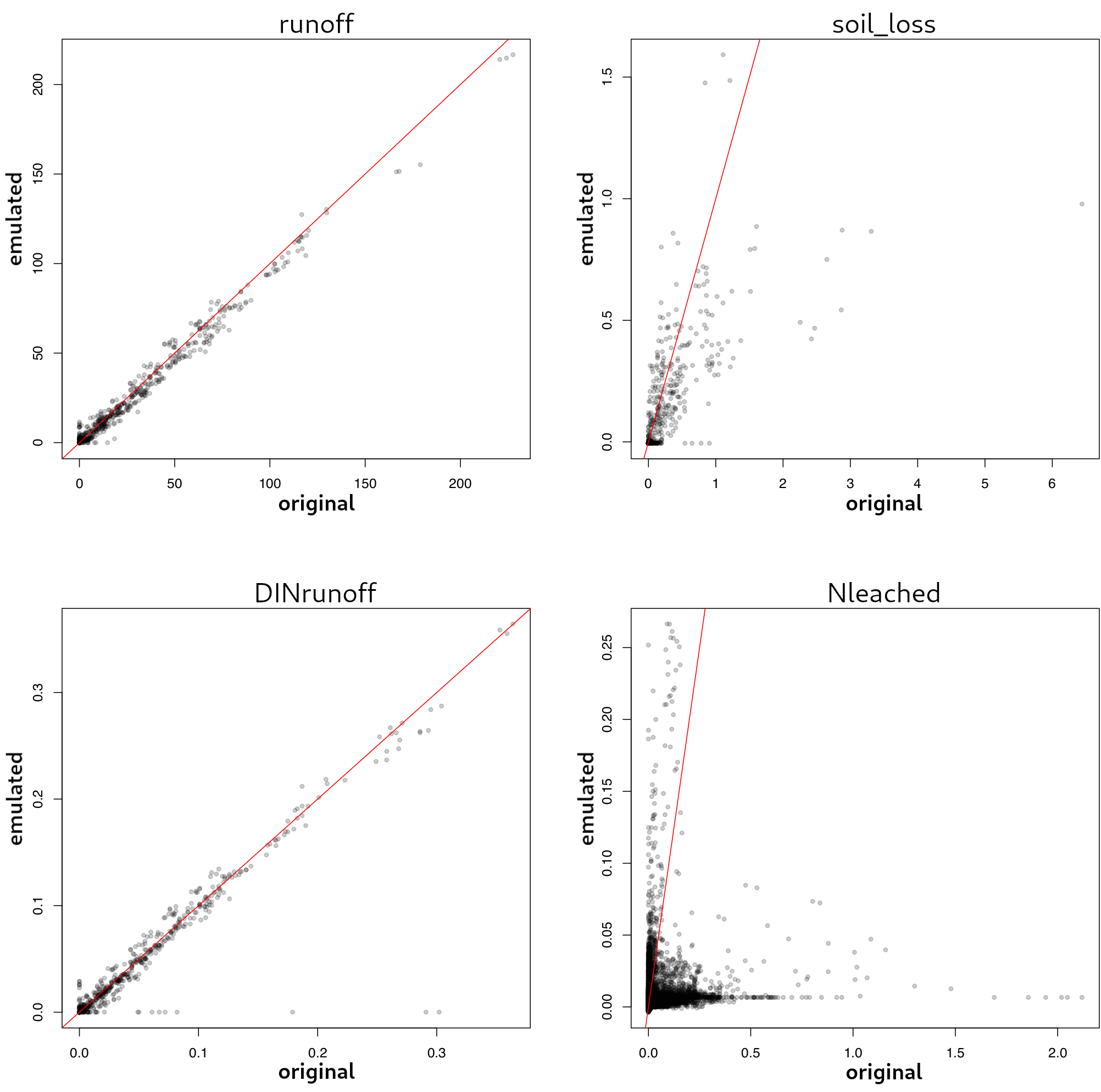}\\
  \end{minipage}
  \begin{minipage}{1\textwidth}
  ~\\
  \hrule
  ~\\
  \end{minipage}
  \begin{minipage}{1\textwidth}
  \centering
  \footnotesize
  {\bf Emulation via GRU-FFNN}\\
  \includegraphics[height=0.65\textwidth,width=0.78\textwidth]{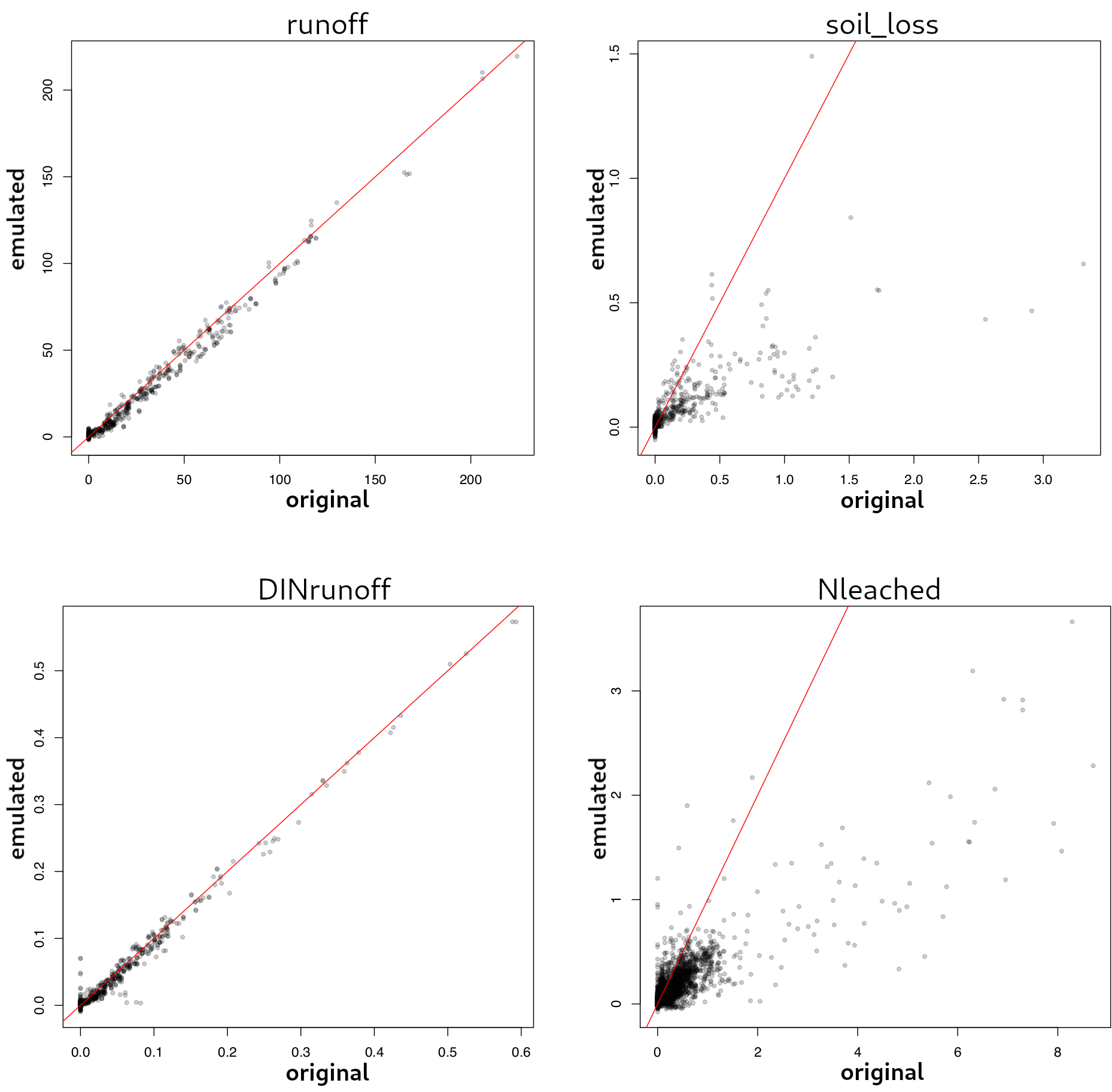}
  \end{minipage}
\end{minipage}
\caption
  {
  Scatter plots of APSIM outputs (original) versus emulated outputs for a cluster of model runs with higher than 0.95 correlation.
  Red lines show equality.
  Top part: emulation using FFNN with 3 layers containing 128, 64 and 32 units, respectively.
  Bottom part: emulation using GRU-FFNN with 3 layers and 128 units per layer;
  corresponding error measures are in Table~\ref{em:metrics}.
  Plots for {\it soil\_loss} and {\it Nleached} have varying scales to reveal details.
  }
\label{em:gru_and_ffnn_emulators_95}
\vspace{-3ex}
\end{figure}

We first examined the performance of emulators constructed on clustered data whose minimum correlation
across the four output variables was greater than 0.95.
Figure~\ref{em:gru_and_ffnn_emulators_95} visually demonstrates the performance of the FFNN and GRU-FFNN emulators constructed for model runs on one soil type,
using clustered data from 3 management classes.
Overall, good results were obtained for emulating {\it runoff} and {\it DINrunoff},
but not for {\it soil\_loss} and {\it Nleached}.
In general we found the GRU-FFNN emulator with a moderate number of hidden layers and units per layer (3~layers, 128~units per layer)
to be most effective. 
The FFNN emulator predicts the {\it Nleached} variable comparatively poorly.

\begin{table}[!tp]
\small
\centering
\begin{tabular}{|c||c|c|c|c|}
\hline
{\bf Min. Corr.} &  {\bf Output Variable} & {\bf MSE} & {\bf MAE} & {\bf Bias}  \\
\bf{Threshold} &  &  &  &    \\  
\hline
0.95  &  {\it runoff}      & 0.612                 & 0.147                 &  -0.0428                \\
      &  {\it soil\_loss}  & $2.30 \times 10^{-3}$ & $4.21 \times 10^{-3}$ &  $-3.18 \times 10^{-3}$ \\
      &  {\it DINrunoff}   & $3.68 \times 10^{-6}$ & $3.64 \times 10^{-4}$ &  $-1.68 \times 10^{-4}$ \\
      &  {\it Nleached}    & $4.12 \times 10^{-2}$ & 0.0404                &  -0.0251                \\
\hline
0.90  &  {\it runoff}      & 0.637                 & 0.135                 & -0.0446                \\
      &  {\it soil\_loss}  & $4.19 \times 10^{-3}$ & $5.82 \times 10^{-3}$ & $-2.01 \times 10^{-3}$ \\
      &  {\it DINrunoff}   & $3.11 \times 10^{-5}$ & $4.70 \times 10^{-4}$ & $-1.38 \times 10^{-4}$ \\
      &  {\it Nleached}    & $2.73 \times 10^{-3}$ & 0.0143                & $-4.54 \times 10^{-3}$ \\
\hline
0.85  &  {\it runoff}      & 0.399                 & 0.120                 & 0.0205                 \\
      &  {\it soil\_loss}  & $5.45 \times 10^{-3}$ & $5.88 \times 10^{-3}$ & $-4.11 \times 10^{-3}$ \\
      &  {\it DINrunoff}   & $3.64 \times 10^{-5}$ & $4.16 \times 10^{-4}$ & $-1.35 \times 10^{-4}$ \\
      &  {\it Nleached}    & $4.82 \times 10^{-3}$ & 0.0197                & $-7.39 \times 10^{-3}$ \\
\hline
\end{tabular}
\caption
  {
  Accuracy metrics for 3 layer (128 nodes per layer) GRU-FFNN emulators built using clustered data obtained from various minimum correlation thresholds.
  MSE: mean squared error. MAE: mean absolute error.
  }
\label{em:metrics}
\end{table}

Metrics of emulation accuracy for the GRU-FFNN emulator are listed in Table~\ref{em:metrics}.
While the error metrics and biases tend to be small,
Figure~\ref{em:gru_and_ffnn_emulators_95} highlights that some outputs were not emulated well at higher values,
and that the small number of these samples (containing such higher values) contribute little to the overall metrics.

\newpage

With the aim of improving the performance of the emulator,
we evaluated lowering the minimum correlation threshold from 0.95 to 0.90 and 0.85,
to establish whether more training data might improve the emulation of {\it Nleached} and {\it soil\_loss}.
This approach appeared to improve the accuracy of predictions of the {\it Nleached} variable,
as shown in Table~\ref{em:metrics}.
However, in lowering the minimum correlation threshold,
we see an undesirable deterioration in the emulation accuracy of the {\it DINrunoff} variable (previously well emulated),
as well as a slight~deterioration in the emulation accuracy of the {\it soil\_loss} variable (\ie contrary to the desired outcome).

In building emulators from the existing APSIM simulation outputs,
the apparent dilemma is that we need more data to train a better model,
but in reducing the minimum correlation threshold to admit more data,
we are introducing data from other management scenarios that have systematic differences in their dynamics.
This suggests that more data (\ie longer time series) is needed from within the same system dynamics (\ie the same management scenarios).

\section{Discussion}
\label{sec:conclusion}
\vspace{-1ex}

The obtained results suggest that the opportunistic re-use of existing datasets
for developing model emulators of APSIM dynamics is limited to a subset of output variables.
Despite large volumes of data in the given dataset of model runs,
the dataset has several limitations: 
sparsity in many of the variables and temporal similarities between various meteorology settings (\eg rain patterns).
This suggests that there is limited (and potentially redundant) information in the dataset,
which may have hindered training of an accurate emulator.

We hypothesise that the differences in accuracy when predicting the four variables
stem from the relative complexity of the relationship between input variables and the outputs.
As such, a greater volume of data may be required to adequately learn more complex model dynamics.
Interestingly, {\it soil\_loss} was not emulated well (particularly at higher values),
despite the fact that {\it runoff} was emulated well and that this would be one of the primary drivers of erosion.
This highlights that there are other processes involved in driving soil loss
(\eg interactions between {\it runoff} and the developmental stage of the cane)
that had not been adequately represented in the training dataset.

To develop emulators capable of adequately replicating a wide range of outputs with more complex dynamics,
we suggest that a ``design of computer experiments'' approach \cite{sacks_1989} should precede the development of emulators,
and be used to generate a richer dataset from which the dynamics can be learned.
In the context of emulating APSIM models,
this can involve sampling a wide variety of combinations of meteorological variables to span the range of all possible inputs to the model.

In particular, we suggest that the next step is to create training data for emulators
by feeding a single, very long (say 1000 year), synthetic, randomly generated meteorology file through each of the APSIM scenario files
(\ie~each combination of farm management conditions and soil type),
rather than using meteorology data from a relatively small number of neighbouring regions.

\section*{Acknowledgements}
\vspace{-2ex}

\begin{small}
We would like to thank the Science Division within the Queensland Government Department of Environment \& Science
for providing the P2R dataset of APSIM model runs.
We would also like to thank Mark Silburn, David Waters, Robin Ellis, Paul Lawrence and Dan Gladish for fruitful discussions.
The PyArmadillo library was used for data preparation~\cite{Rumengan_JOSS_2021}.
\end{small}

\newpage

\righthyphenmin=100
\bibliographystyle{ieee}
\small
\bibliography{references}
\end{document}